\PassOptionsToPackage{table}{xcolor}

\documentclass{article}
\usepackage[most]{tcolorbox}
\usepackage{listings}
\usepackage[table]{xcolor}
\usepackage{capt-of}
\definecolor{promptcolor}{HTML}{EAF3FF}
\definecolor{promptcolorheader}{HTML}{8DB7E8}
\definecolor{codeback}{HTML}{F7FAFF}

\newtcolorbox{promptbox}[1][]{%
    enhanced,
    breakable,
    colback=promptcolor!50,
    colframe=promptcolorheader,
    colbacktitle=promptcolorheader,
    coltitle=black,
    boxrule=0.5pt,
    arc=2pt,
    outer arc=2pt,
    width=\columnwidth,
    title={\footnotesize\bfseries #1},
    fonttitle=\footnotesize\bfseries,
    left=3pt,
    right=3pt,
    top=3pt,
    bottom=3pt,
    before skip=6pt,
    after skip=6pt
}

\usepackage{vigor_arxiv,times}

\usepackage{amsmath,amsfonts,bm}

\def\eqref#1{equation~\ref{#1}}
\def\1{\bm{1}}

\DeclareMathAlphabet{\mathsfit}{\encodingdefault}{\sfdefault}{m}{sl}
\SetMathAlphabet{\mathsfit}{bold}{\encodingdefault}{\sfdefault}{bx}{n}

\usepackage{graphicx}
\usepackage{float}
\usepackage{xcolor}
\usepackage{makecell}
\usepackage{hyperref}
\usepackage{url}
\usepackage{wrapfig}
\usepackage{amssymb}
\usepackage{booktabs}

\definecolor{bestblue}{HTML}{D9EEF7}
\definecolor{secondpink}{HTML}{F7E1ED}

\title{VIGOR: Verifiable Industrial Geometry Optimization and Repair in Scene Programs}

\author{
Fuqiang Wang$^{2}$,
Zheng Guo$^{2}$,
Xiaoming Wu$^{2}$,
Hanlin Kong$^{2}$, \\
Nayu Liu$^{3}$,
Chuanting Zhang$^{4}$,
Daojun Liang$^{5}$,
Jingxuan Wei$^{1,2,6,*}$ \\[4pt]
$^{1}$University of Chinese Academy of Sciences, Beijing, China \\
$^{2}$Qilu University of Technology, Jinan, China \\
$^{3}$Tianjin University, Tianjin, China \\
$^{4}$Shandong University, Jinan, China \\
$^{5}$Qilu University of Technology (Shandong Academy of Sciences), Jinan, China \\
$^{6}$Shandong Computer Science Center, Jinan, China \\
$^{*}$Corresponding author: Jingxuan Wei , \texttt{weijingxuan20@mails.ucas.edu.cn} 
}

\iclrfinalcopy

\begin{document}
\raggedbottom
\maketitle

\begin{abstract}
Large language models (LLMs) increasingly generate structured layouts and executable scene programs from natural-language instructions. However, \textbf{executability does not imply engineering feasibility}: an industrial scene program may execute successfully while its final state still violates coupled boundary, collision, clearance, aisle, or structural constraints. We identify this mismatch as the \textbf{\emph{executability--engineering-feasibility gap}}, where locally plausible placements can remain globally infeasible because corrections to one device may introduce violations elsewhere. To make this gap measurable in a controlled setting, we construct an execution-grounded industrial scene-program dataset with \textbf{1,749 constructible scenarios}, complete executable programs, and \textbf{94 final-state geometric checks per scenario}. We further propose \textbf{VIGOR}, which combines supervised scene-program generation with execution-grounded verification and staged geometric refinement. Supervised fine-tuning establishes complete program generation, \textbf{local projection} resolves independently correctable violations, and \textbf{joint conflict-graph repair} coordinates residual multi-object conflicts while preserving structural relations. Experiments show that VIGOR achieves \textbf{96.86\% Complete Feasibility} and an \textbf{Overall score of 91.20}, outperforming the strongest evaluated baseline by \textbf{10.42 points in Overall}. These results show that, under controlled industrial layout conditions, reliable scene-program generation requires not only executable programs, but also explicit verification and coordinated resolution of coupled engineering constraints.
\end{abstract}

\section{Introduction}

\begin{figure}[t]
    \centering
    \includegraphics[width=\linewidth]{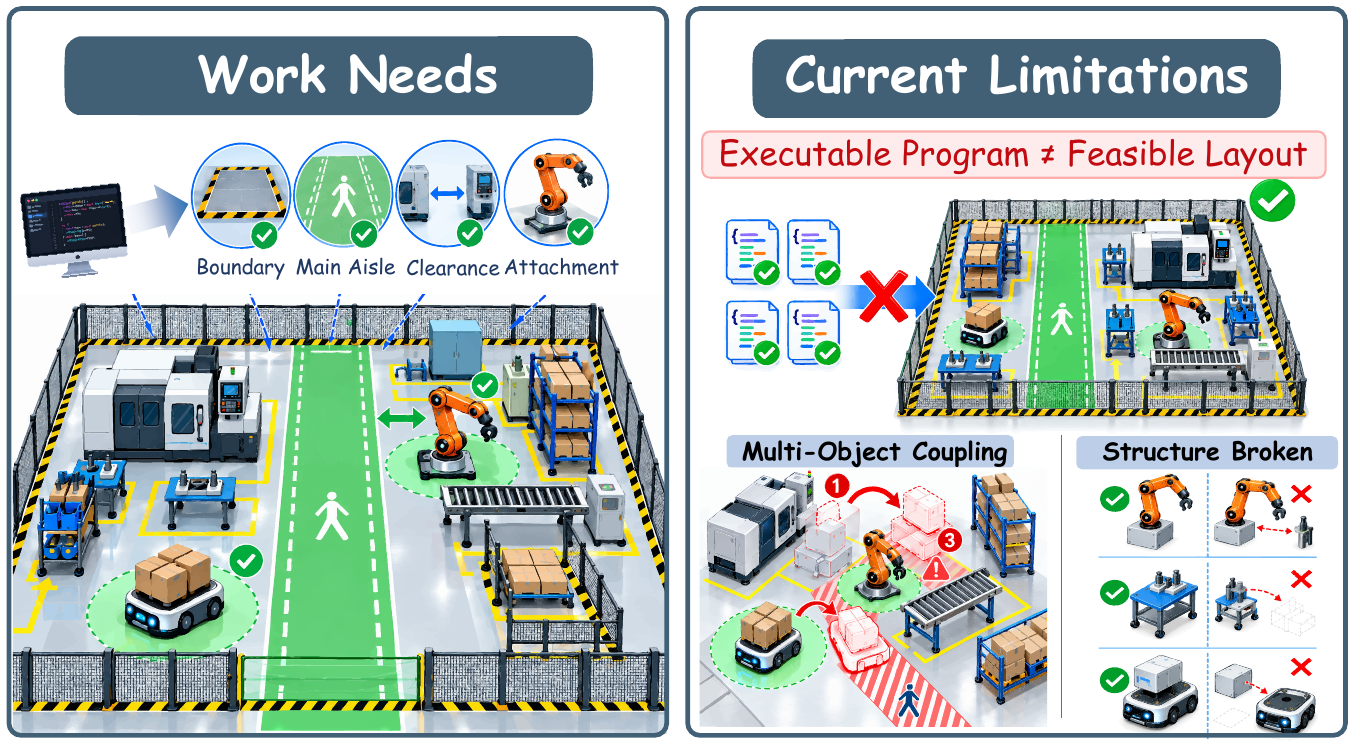}
    \caption{Industrial scene-program generation exposes an engineering-feasibility gap hidden by successful execution. A valid industrial layout must simultaneously satisfy boundary, aisle, clearance, and attachment constraints, while executable programs may still contain coupled multi-object conflicts or broken structural relations that invalidate the final scene.}
    \label{fig:motivation}
\end{figure}

Language-driven scene generation increasingly turns natural-language requirements into structured layouts and executable scene programs. Recent systems have advanced from compositional layout synthesis toward programmatic and industrial scene construction~\citep{feng2023layoutgpt,lin2024instructscene,hu2024scenecraft,xia2025scenegenagent,zhang2025scenelanguage}. However, successful execution does not guarantee that the resulting layout satisfies its engineering constraints. This leaves a basic capability boundary unresolved: can a model generate industrial scene programs that are \textbf{not only executable, but also engineering-feasible}?

We investigate this boundary through industrial scene-program generation. Given a natural-language requirement and explicit engineering constraints, the model must generate a complete structured program that instantiates and arranges the required devices while preserving structural relations. We intentionally adopt a \textbf{controlled industrial setting} with a fixed asset inventory and program schema, allowing us to systematically isolate the central question of whether executable scene programs can satisfy coupled engineering constraints under varying spatial conditions. As illustrated in Figure~\ref{fig:motivation}, this setting changes the success criterion from program execution to \textbf{complete engineering feasibility of the final scene state}. Moreover, \textbf{engineering constraints are coupled through shared device transforms}: adjusting one device to satisfy a local constraint may create new collisions, clearance violations, or aisle conflicts elsewhere.

We refer to this mismatch as the \textbf{\emph{executability--engineering-feasibility gap}}. It lies at the intersection of executable scene, constraint-aware layout reasoning, and execution-guided repair. Existing work increasingly produces structured or executable scenes~\citep{hu2024scenecraft,xia2025scenegenagent,zhang2025scenelanguage,wang2026scenecode}, improves spatial and multi-object validity~\citep{wang2019planit,dupty2024factorgraphs,sun2025layoutvlm,li2026constraintir}, or uses execution feedback and geometric validation to refine generated artifacts~\citep{zhong2024ldb,gumin2025proceduralsceneprograms,barkley2026cadsmith}. Yet complete post-execution feasibility under all task-specified engineering constraints remains insufficiently characterized. To make this gap measurable, we construct an execution-grounded dataset with \textbf{1,749 constructible scenarios}. Each contains 11 required instances and a complete 25-operation \texttt{scene\_program.v1}, whose final state is evaluated for structural consistency and \textbf{94 geometric checks} covering boundaries, collisions, clearances, and the main aisle.

We further propose \textbf{VIGOR}, which couples supervised scene-program generation with execution-grounded verification and staged geometric refinement. Supervised fine-tuning first learns to generate complete programs from task requirements and engineering constraints. Since token-level imitation does not directly optimize final-state feasibility, executable outputs may retain hard-constraint violations. VIGOR therefore verifies the executed state, applies \textbf{local refinement} to independently correctable violations, and invokes \textbf{joint refinement} for residual coupled conflicts requiring coordinated multi-object adjustments. The refined program is then re-executed and re-verified.

Experiments reveal a pronounced \textbf{\emph{executability--engineering-feasibility gap}}: strong models can achieve high geometric quality while still failing complete feasibility. Ablations show that supervised fine-tuning establishes strong structural and geometric priors, \textbf{local refinement resolves independent violations}, and \textbf{joint refinement addresses residual coupled conflicts}. Together, these components yield the strongest complete-feasibility performance.

Our main contributions are as follows:
\begin{itemize}
    \item We formalize \textbf{complete engineering feasibility} for industrial scene-program generation and characterize an \emph{executability--engineering-feasibility gap}, where successful execution does not guarantee satisfaction of all task-specified engineering constraints.

    \item We construct a \textbf{controlled execution-grounded benchmark for industrial scene-program generation} with 1,749 constructible scenarios, complete structured programs, and final-state verification of structural consistency and geometric feasibility under systematically varied engineering conditions.

    \item We propose \textbf{VIGOR}, combining supervised scene-program generation with execution-grounded verification and complementary \textbf{local--joint geometric refinement}. Experiments show that VIGOR substantially improves complete engineering feasibility over general models and representative code and layout generation methods.
\end{itemize}

\section{Related Work}
\label{sec:related_work}

\subsection{Language-Driven Executable Scene Generation}

Language-driven scene generation translates natural-language descriptions into structured layouts, 3D environments, or executable scene programs. Prior work has advanced from object layout and semantic relation modeling to programmatic scene construction and executable generation \citep{feng2023layoutgpt,lin2024instructscene,hu2024scenecraft,khan2024text2cad,zhang2025scenelanguage}. Recent systems further extend this paradigm toward industrial, physically validated, and simulation-ready scenes \citep{xia2025scenegenagent,xia2026sage,wang2026scenecode,xiao2026roomwright}. Our work studies complete executable scene generation from natural-language industrial requirements, emphasizing the spatial and structural feasibility of the final executed state.

\subsection{Scene Constraint Satisfaction and Complete Feasibility}

Constraint-aware scene generation incorporates spatial, physical, and relational conditions to improve geometric and functional feasibility \citep{wang2019planit,yang2024holodeck,yang2024physcene,raistrick2024infinigenindoors}. Subsequent methods model interacting constraints through structured representations, optimization, and global validation to reduce collisions, boundary violations, and other layout inconsistencies \citep{dupty2024factorgraphs,sun2025layoutvlm,wang2026function2scene,li2026constraintir,huang2026ggvo,bisht2022g2plan}. Our work instead focuses on complete feasibility: whether the final executed scene simultaneously satisfies all task-specified spatial and structural constraints.

\subsection{Execution-Guided Scene Generation and Scene Legalization}

Execution and verification feedback have long been used in program synthesis to refine generated programs through tests, runtime states, or counterexamples \citep{solarlezama2006sketch,nguyen2013semfix,mechtaev2015directfix,mechtaev2016angelix,le2022coderl,zhong2024ldb}. Related ideas have been adopted in scene and CAD generation through execution feedback, geometric verification, program search, and layout legalization \citep{hu2024scenecraft,xia2025scenegenagent,barkley2026cadsmith,gumin2025proceduralsceneprograms,spindler2008abacus,vonessen2013repairwithoutregret}. Our work integrates legalization into language-to-scene generation, correcting residual spatial violations while preserving program structure, instance identities, hierarchy, and attachment consistency.

\section{Method}

\begin{figure}[t]
    \centering
    \includegraphics[width=1.02\linewidth]{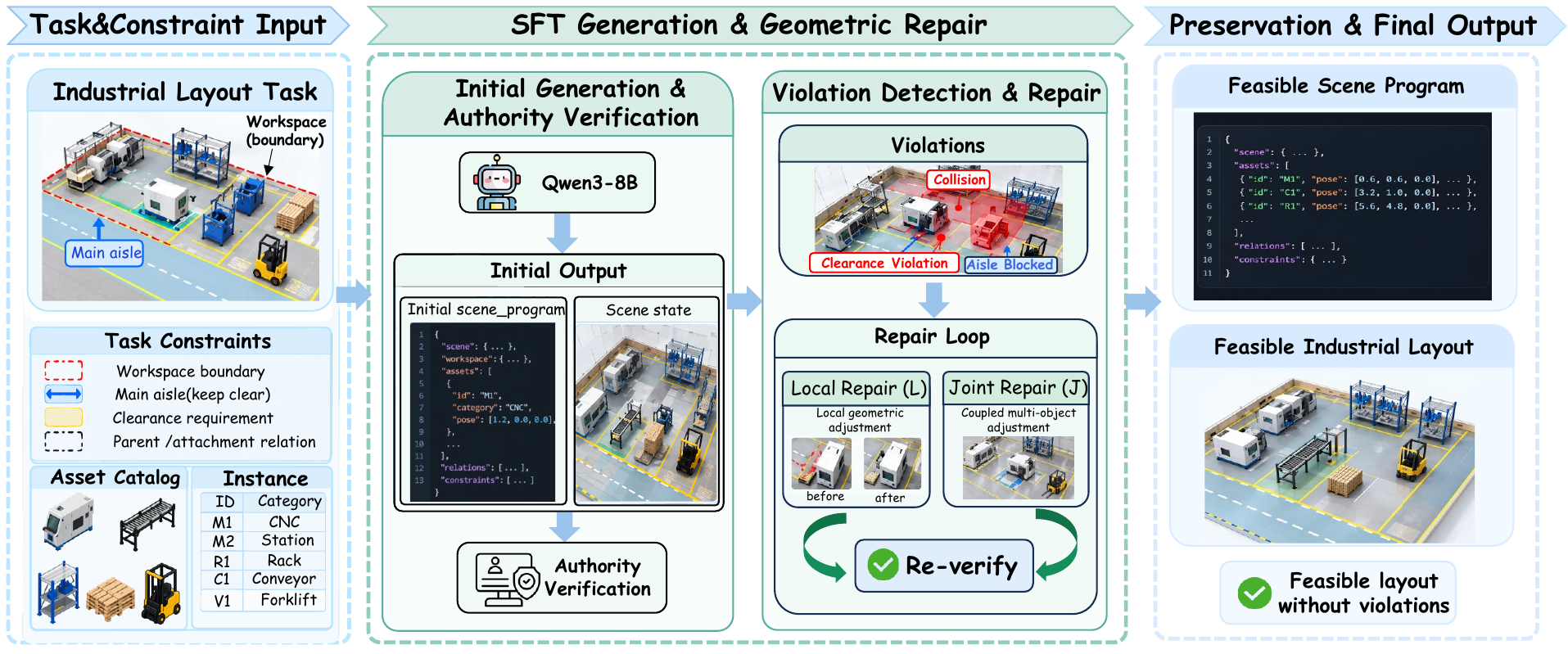}
    \caption{Overview of VIGOR. SFT generates the initial scene program, verification detects residual violations, and local--joint repair produces a feasible layout.}
    \label{fig:overall}
\end{figure}

\subsection{Problem Formulation}

We study executable industrial scene-program generation. Given context
$X$ containing an instruction, constraints, asset geometry, and
instance/hierarchy contracts, the model generates a program
\begin{equation}
P=(a_1,\ldots,a_L),\qquad
a_\ell=(\kappa_\ell,\xi_\ell),\qquad
S_\ell=\mathcal E(S_{\ell-1},a_\ell),
\end{equation}
where $\mathcal E$ is the frozen state engine,
$\kappa_\ell\in\{\texttt{create},\texttt{transform},
\texttt{parent},\texttt{save}\}$ is the operation type, and
$\xi_\ell$ contains its instance identifiers and typed parameters.

Successful execution alone is insufficient. Let $\mathbf t(P)$
denote the executed transforms and $\mathcal H(X)$
workspace, boundary, collision, clearance, aisle, and structural
constraints. Complete feasibility is
\begin{equation}
\mathrm{Feasible}(P\mid X)=
\begin{cases}
\displaystyle\prod_{h\in\mathcal H(X)}
h(\mathbf t(P);X),
& \mathrm{Exec}(P)=1,\\
0, & \mathrm{otherwise},
\end{cases}
\end{equation}
where the predicates are evaluated on the final reloaded scene,
including its identities and hierarchy.

These constraints are coupled through shared transforms. For
constraint $h$, let $m_h(\mathbf t_{\mathcal I(h)};X)$ denote its
signed satisfaction margin over the involved devices
$\mathcal I(h)$. A correction $\Delta\mathbf t$ induces
\begin{equation}
\Delta m_h
=
m_h\!\left((\mathbf t+\Delta\mathbf t)_{\mathcal I(h)};X\right)
-
m_h\!\left(\mathbf t_{\mathcal I(h)};X\right),
\quad
\mathcal I(h)\cap
\operatorname{supp}(\Delta\mathbf t)\neq\varnothing.
\end{equation}
Thus, improving one constraint may inadvertently degrade others, motivating iterative local
correction followed by joint refinement of residual conflicts.

\subsection{VIGOR: Generation with Local--Joint Refinement}

As shown in Figure~\ref{fig:overall}, VIGOR combines scene-program
generation, execution-grounded verification, and staged geometric
refinement. The generator produces $P_0=f_\theta(X)$, which is
executed and verified against the public constraints. The verifier
returns $\mathcal V(P,X)$, containing violated constraint atoms,
their involved devices, and violation deficits. For an executable
$P_0$, VIGOR proceeds as
\begin{equation}
\begin{aligned}
P_{\mathrm{loc}}
&=
\begin{cases}
P_0,
& \mathrm{Feasible}(P_0\mid X)=1,\\
\mathcal L(P_0,X;\mathcal V(P_0,X)),
& \mathrm{otherwise},
\end{cases}
\\
P_{\mathrm{F}}
&=
\begin{cases}
P_{\mathrm{loc}},
& \mathrm{Feasible}(P_{\mathrm{loc}}\mid X)=1,\\
\mathcal J(P_{\mathrm{loc}},X;
           \mathcal V(P_{\mathrm{loc}},X)),
& \mathrm{otherwise}.
\end{cases}
\end{aligned}
\label{eq:overall_refinement}
\end{equation}

Local refinement $\mathcal L$ handles independently correctable
violations, while joint refinement $\mathcal J$ coordinates residual
multi-object conflicts. Candidate programs are re-executed and
verified, while preserving the program structure, instance identities,
hierarchy, and attachment relations. Non-executable outputs bypass
refinement and remain failed.

\subsection{Supervised Scene-Program Generation}

Given training pairs $(X,P^*)$, where $P^*$ is a reference
\texttt{scene\_program.v1}, we fine-tune Qwen3-VL-8B-Instruct~\citep{bai2025qwen3vl} with
LoRA. Writing $P^*$ as $(p_1^*,\ldots,p_M^*)$, we minimize
\begin{equation}
\mathcal L_{\mathrm{SFT}}
=
-\sum_{(X,P^*)\in\mathcal D_{\mathrm{train}}}
\sum_{m=1}^{M}
\log p_\theta(p_m^*\mid X,p_{<m}^*).
\label{eq:sft_objective}
\end{equation}
The supervision covers complete programs, including device creation,
transform assignment, hierarchy construction, and scene saving.The complete \texttt{scene\_program.v1} interface and model-visible generation template are provided in Appendix~\ref{app:program_specification}.

At inference time, a single greedy decoding pass produces $P_0$.
Since token-level imitation does not directly optimize final-state
feasibility, the generated program is executed and verified, and any
residual violations are passed to geometric refinement.

\subsection{Local--Joint Geometric Refinement}
\label{sec:geometric_refinement}

Geometric refinement operates on the executed scene through local
projection followed by joint conflict resolution. Local candidates
$\mathcal C_{\mathcal L}(P)$ apply boundary projections, aisle
push-outs, and clearance corrections. Their admissibility is
determined after re-execution:
\begin{equation}
\begin{aligned}
\mathcal A_{\mathcal L}(P)=
\bigl\{P'\in\mathcal C_{\mathcal L}(P):\;&
\mathrm{Valid}(P'),\
\mathcal V(P',X)\subseteq\mathcal V(P,X),\\
&n_{\mathcal T}(P')<n_{\mathcal T}(P)
\lor D_{\mathcal T}(P')<D_{\mathcal T}(P),\\
&\mathcal V(P',X)=\mathcal V(P,X)
\Rightarrow s(P')\ge s(P)\bigr\},
\end{aligned}
\label{eq:local_admissibility}
\end{equation}
where $\mathcal T$ denotes the targeted constraint atoms,
$n_{\mathcal T}$ counts violations, $D_{\mathcal T}$ sums their
deficits, and $s$ is mean functional quality. $\mathrm{Valid}(P')$
requires successful execution and structural correctness. Among
admissible candidates, local refinement favors fewer modified devices
and smaller displacement.

Residual violations induce a conflict graph whose connected
components identify coupled device groups. For a valid search state
$z=(P,\mathcal M,D)$, joint refinement uses the lexicographic objective
\begin{equation}
\boldsymbol\Phi(z)=
\left(
|\mathcal V|,
\sum_{h\in\mathcal V}\bar d_h,
\max_{h\in\mathcal V}\bar d_h,
-s(P),|\mathcal M|,D,\sigma(P)
\right),
\label{eq:joint_objective}
\end{equation}
where $\bar d_h$ is the normalized violation deficit,
$\mathcal M$ records modified devices, $D$ accumulates displacement,
and $\sigma(P)$ provides deterministic tie-breaking.

Let $\mathcal N(z)$ contain all valid successors that preserve satisfied
parent-link and attachment constraints. Beam search iteratively updates
\begin{equation}
\mathcal B_{d+1}
=
\operatorname{TopB}_{\mathrm{lex}\,\boldsymbol\Phi}
\left(
\bigcup_{z\in\mathcal B_d}\mathcal N(z)
\right).
\label{eq:joint_refinement}
\end{equation}
Search terminates upon feasibility or budget exhaustion, and the best
state is retained only when it improves the input objective and
preserves attachment consistency.

\begin{figure}[H]
    \centering
    \includegraphics[width=\linewidth]{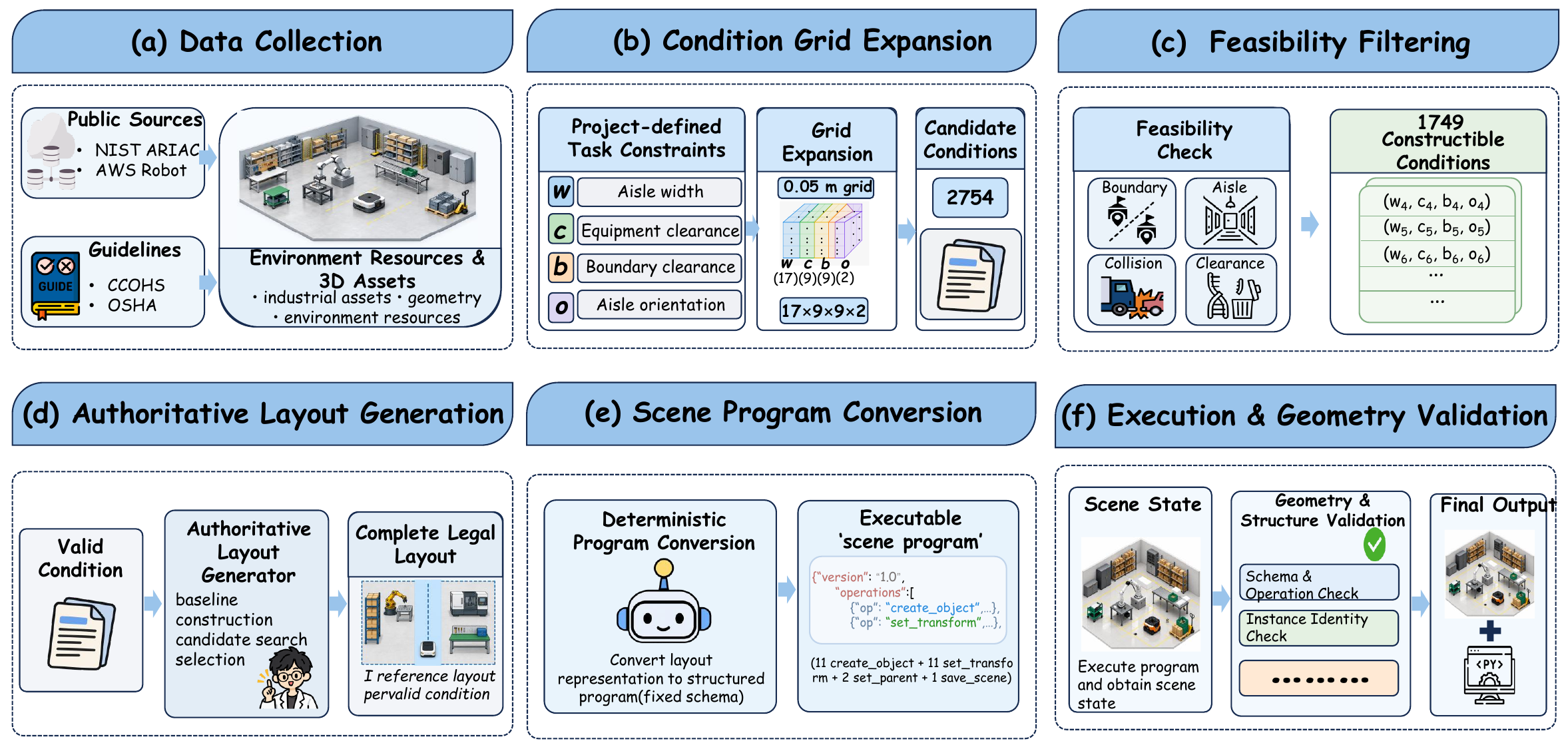}
    \caption{Construction pipeline of the VIGOR dataset. Public assets and engineering constraints are expanded into candidate conditions, filtered for constructibility, converted into executable scene programs, and finally executed and validated.}
    \label{fig:data_construction}
\end{figure}

\section{Dataset}

\label{sec:dataset}
\subsection{Dataset Construction}

\paragraph{Asset and Task Condition Construction.}
Candidate assets and environments are collected from NIST ARIAC, Poly Haven, and AWS RoboMaker Small Warehouse. Since these resources cannot directly form a unified industrial layout task, we standardize units, coordinate systems, pivots, and local AABBs, and unify instance identities, parent--child relations, and the scene-program interface. Based on these standardized resources, we construct a candidate condition space over main-aisle width, device clearance, boundary clearance, and aisle orientation, providing unified inputs for subsequent constructibility screening and reference layout generation. The complete condition grid and constructibility-screening protocol are provided in Appendix~\ref{app:task_construction}.

\paragraph{Constructibility Screening and Reference Layout Generation.}
Candidate conditions are first processed by a predefined layout generation and validation pipeline to remove those for which a complete valid layout cannot be constructed under the current construction procedure. For each retained condition, a unified layout generator and deterministic search procedure are used to produce candidate reference layouts, from which the final layout is selected through a consistent validation and scoring process. This stage yields a complete reference layout aligned with each task condition, serving as the intermediate representation for subsequent structured scene-program conversion.

\paragraph{Structured Scene-Program Conversion.}
Each reference layout is then mapped to a unified \texttt{scene\_program.v1} representation. Object identities, spatial transformations, and parent--child relations are encoded as object-creation, transformation-setting, and hierarchy-setting operations, respectively, and organized according to a consistent program order. This conversion only structures the established reference layout without re-searching or modifying its spatial configuration. As a result, each reference layout is converted into a complete executable scene program for subsequent execution and state validation. The complete program interface and model-visible generation template are provided in Appendix~\ref{app:program_specification}.

\paragraph{Execution Validation and Final Packaging.}
Each complete scene program is executed by a Python state engine, and the final scene state is obtained after saving and reloading. The executed state is then consistently validated for instance completeness, parent--child and attachment relations, as well as spatial conditions including boundaries, collisions, clearances, and the main aisle. The validated task input and scene program are finally packaged as a dataset sample, ensuring that each supervision target has both a standardized program structure and an execution-verified complete scene state. Detailed schema, execution, save--reload, structural, and geometric validation procedures are provided in Appendix~\ref{app:dataset_quality}.

\subsection{Dataset Analysis}

\begin{wrapfigure}[16]{r}{0.35\linewidth}
    \centering
    \vspace{-8pt}
    \includegraphics[width=\linewidth]{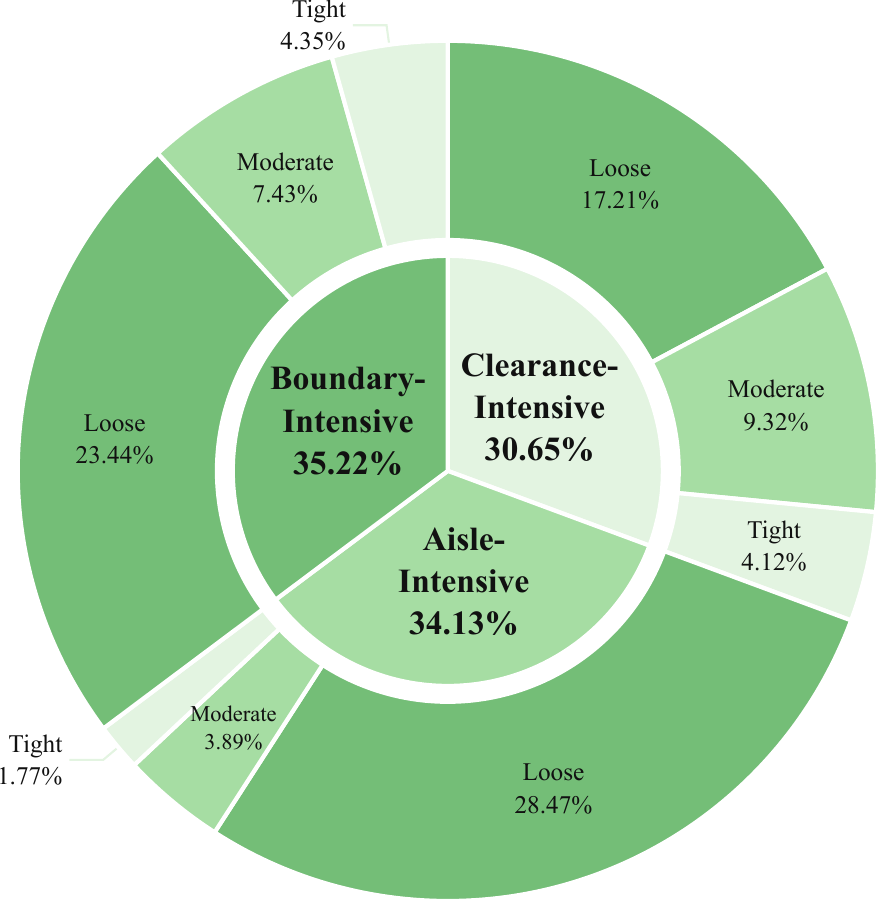}
    \caption{Constraint-profile composition of the VIGOR dataset.}
    \label{fig:data_analysis}
    \vspace{-8pt}
\end{wrapfigure}
Figure~\ref{fig:data_analysis} and Table~\ref{tab:dataset_statistics} summarize the engineering-constraint composition and structured generation scale of the dataset. From 2,754 candidate conditions, we obtain 1,749 constructible scenarios, which are split into 1,399 training samples, 175 interpolation test samples, and 175 compositional test samples. To characterize the variation in engineering conditions within the final dataset, we perform a normalized post-hoc grouping based on main-aisle width, equipment clearance, and boundary clearance, resulting in three similarly sized constraint profiles. Each profile spans Loose, Moderate, and Tight feasibility margins, indicating that constraint composition and feasibility tightness constitute two complementary dimensions of dataset variation.

On the structured-generation side, each scenario contains 11 instances and a complete, fixed-length 25-operation \texttt{scene\_program.v1} program, covering object creation, spatial transformation, parent--child relations, and scene saving. After program execution and save/reload, each scenario is further subjected to 94 geometric checks covering boundary constraints, collisions, equipment clearances, and main-aisle constraints. All 1,749 reference programs pass the complete validation pipeline.

\begin{table}[t]
\centering
\caption{Key statistics of the industrial scene program dataset.}
\label{tab:dataset_statistics}

\small
\renewcommand{\arraystretch}{1.12}

\begin{tabular*}{\linewidth}{
@{\extracolsep{\fill}}
l r l r
@{}
}
\toprule

\multicolumn{2}{c}{\textbf{Dataset statistics}} &
\multicolumn{2}{c}{\textbf{Generation statistics}} \\

\cmidrule(lr){1-2}
\cmidrule(lr){3-4}

\textbf{Statistic} &
\textbf{Count} &
\textbf{Statistic} &
\textbf{Count} \\

\midrule

Candidate conditions
& 2,754
& Scene instances
& 19,239 \\

Final scenarios
& 1,749
& Scene programs
& 1,749 \\

Train / I / C
& 1,399 / 175 / 175
& Program operations
& 43,725 \\

Reference layouts
& 1,749
& Geometric checks
& 164,406 \\

Validated scenarios
& 1,749
& Verified programs
& 1,749 \\

\bottomrule
\end{tabular*}

\end{table}

\section{Experiments}

\subsection{Experimental Setup}

We use Qwen3-VL-8B-Instruct~\citep{bai2025qwen3vl} as the base model and apply LoRA for supervised fine-tuning (SFT). The training set contains 1,399 samples, and the model is trained for 3 epochs, corresponding to 4,197 sample exposures and 525 optimization steps. At test time, we adopt single-pass greedy decoding, where each task produces exactly one complete \texttt{scene\_program.v1} program without retries or multi-candidate sampling.

We systematically compare VIGOR against three distinct groups of methods. Open-weight models include GLM-4-9B-0414~\citep{zai2025glm49b0414}, Qwen3-14B~\citep{yang2025qwen3}, Gemma-3-12B-IT~\citep{gemmateam2025gemma3}, DeepSeek-Coder-V2-Lite-Instruct~\citep{deepseekai2024deepseekcoderv2}, Granite-3.3-8B-Instruct~\citep{ibm2025granite33}, Qwen3.5-9B~\citep{qwen2026qwen35}, and Ministral-3-8B-Instruct-2512~\citep{liu2026ministral3}. Closed-source/API models include GPT-5.5~\citep{openai2026gpt55}, Claude Sonnet 5~\citep{anthropic2026claudesonnet5}, DeepSeek-V4.1-Flash~\citep{deepseekai2026deepseekv41flash}, and Grok-4.5~\citep{xai2026grok45}. We further compare against published code and layout generation methods, including OpenCodeReasoning-Nemotron-7B~\citep{ahmad2025opencodereasoning}, CodeRL+~\citep{jiang2026coderl+}, and ReSpace~\citep{bucher2025respace}. This comprehensive benchmark suite covers general open-weight generative models, proprietary large language models, and representative methods for code and structured layout generation. Detailed metric definitions and aggregation rules are provided in Appendix~\ref{app:evaluation_metrics}, and the complete machine-evaluation prompts are provided in Appendix~\ref{app:machine_evaluation_prompts}.

\subsection{Main Results}

Table~\ref{tab:main_results} shows that VIGOR achieves the best Overall score of 91.20, outperforming the strongest baseline, DeepSeek-V4.1-Flash, by 10.42 points (12.9\%). It also obtains the best Bottleneck Geometry, Functional Layout, and Tight-FQ scores, indicating stronger geometric feasibility, functional quality, and robustness under tight constraints. Notably, although DeepSeek-V4.1-Flash reaches comparable GVB and Bottleneck Geometry scores of 0.33 and 98.55, its Functional Layout score is only 42.24, compared with 72.39 for VIGOR. This suggests that VIGOR better converts geometric validity into overall engineering-layout quality.

The results further expose a clear \emph{executability--engineering-feasibility gap}. GPT-5.5 and Claude Sonnet 5 obtain GVB scores of 6.05 and 10.11 and Bottleneck Geometry scores of 77.40 and 55.29, while their Complete Feasibility remains only 11.71 and 0.00, respectively. In contrast, VIGOR reaches 0.30 GVB, 99.51 Bottleneck Geometry, and 96.86 Complete Feasibility. This shows that reliable scene-program generation is bottlenecked not only by program generation, but by simultaneous satisfaction of coupled engineering constraints.

Compared with published code and layout generation methods, VIGOR highlights the limitations of existing approaches. OpenCodeReasoning-Nemotron-7B and CodeRL+ achieve Overall scores of only 0.25 and 1.46, while ReSpace reaches 14.26, far below VIGOR's 91.20. These results indicate that code generation or spatial layout capability alone is insufficient for engineering feasibility.

Overall, reliable industrial scene-program generation requires not only executable structured programs, but also engineering-constraint verification and coordinated resolution of geometric conflicts. Additional process--result and task-capability analyses are provided in Appendices~\ref{app:capability_profiles} and~\ref{app:task_capability}.
\begin{table*}[t]
\centering
\caption{Main results on industrial scene-program generation. Models are grouped into open-weight models, closed-source/API models, published methods, and our method. Best and second-best results are highlighted in blue and pink.}
\label{tab:main_results}

\setlength{\tabcolsep}{2.0pt}
\renewcommand{\arraystretch}{1.05}

\resizebox{\textwidth}{!}{%
\begin{tabular}{@{}l c c c c c c c c c c@{}}
\toprule

\textbf{Method} &
\textbf{GVB}$\downarrow$ &
\textbf{BG}$\uparrow$ &
\textbf{Feas.}$\uparrow$ &
\textbf{Func.}$\uparrow$ &
\textbf{BQ-FQ}$\uparrow$ &
\textbf{Interp-FQ}$\uparrow$ &
\textbf{Comp-FQ}$\uparrow$ &
\textbf{Tight-FQ}$\uparrow$ &
\textbf{Overall}$\uparrow$ &
\textbf{LLMJudge}$\uparrow$ \\

\midrule

\multicolumn{11}{c}{\textit{Open-weight Models}} \\
\midrule

GLM-4-9B-0414~\citep{zai2025glm49b0414}
& 11.30
& 34.96
& 0.00
& \cellcolor{secondpink}\textbf{53.28}
& 24.53
& \cellcolor{secondpink}\textbf{54.56}
& \cellcolor{secondpink}\textbf{52.00}
& \cellcolor{secondpink}\textbf{49.72}
& 27.07
& 1.975 \\

Qwen3-14B~\citep{yang2025qwen3}
& 15.47
& 44.48
& 0.00
& 40.78
& 24.15
& 42.81
& 38.75
& 42.74
& 25.46
& 3.225 \\

Gemma-3-12B-IT~\citep{gemmateam2025gemma3}
& 38.21
& 0.65
& 0.00
& 29.87
& 6.25
& 29.71
& 30.03
& 19.45
& 13.27
& 0.325 \\

DeepSeek-Coder-V2-Lite-Instruct~\citep{deepseekai2024deepseekcoderv2}
& 80.84
& 0.00
& 0.00
& 2.65
& 0.00
& 3.06
& 2.23
& 3.46
& 2.32
& 0.300 \\

Granite-3.3-8B-Instruct~\citep{ibm2025granite33}
& 44.25
& 11.01
& 0.00
& 11.58
& 0.00
& 11.68
& 11.47
& 15.61
& 10.51
& 1.225 \\

Qwen3.5-9B~\citep{qwen2026qwen35}
& 15.97
& 20.32
& 0.00
& 43.02
& 25.33
& 42.80
& 43.24
& 37.30
& 22.50
& 3.275 \\

Ministral-3-8B-Instruct-2512~\citep{liu2026ministral3}
& 18.09
& 64.30
& 0.00
& 34.75
& 26.24
& 36.10
& 33.39
& 39.33
& 26.62
& 3.125 \\

\midrule

\multicolumn{11}{c}{\textit{Closed-source/API Models}} \\
\midrule

GPT-5.5~\citep{openai2026gpt55}
& 6.05
& 77.40
& 11.71
& 37.01
& 18.07
& 37.08
& 36.94
& 38.11
& 55.09
& 3.425 \\

Claude Sonnet 5~\citep{anthropic2026claudesonnet5}
& 10.11
& 55.29
& 0.00
& 45.56
& \cellcolor{secondpink}\textbf{30.89}
& 46.25
& 44.87
& 42.46
& 52.57
& \cellcolor{secondpink}\textbf{3.900} \\

DeepSeek-V4.1-Flash~\citep{deepseekai2026deepseekv41flash}
& \cellcolor{secondpink}\textbf{0.33}
& \cellcolor{secondpink}\textbf{98.55}
& \cellcolor{secondpink}\textbf{84.00}
& 42.24
& 26.39
& 41.98
& 42.50
& 47.00
& \cellcolor{secondpink}\textbf{80.78}
& 3.625 \\

Grok-4.5~\citep{xai2026grok45}
& 25.21
& 41.07
& 0.00
& 26.79
& 6.25
& 26.44
& 27.15
& 27.65
& 23.25
& 2.800 \\

\midrule

\multicolumn{11}{c}{\textit{Published Methods}} \\
\midrule

OpenCodeReasoning-Nemotron-7B~\citep{ahmad2025opencodereasoning}
& 92.91
& 0.28
& 0.00
& 0.00
& 0.00
& 0.00
& 0.00
& 0.00
& 0.25
& 0.125 \\

CodeRL+~\citep{jiang2026coderl+}
& 83.00
& 0.00
& 0.00
& 0.00
& 0.00
& 0.00
& 0.00
& 0.00
& 1.46
& 0.000 \\

ReSpace~\citep{bucher2025respace}
& 14.59
& 29.61
& 0.00
& 0.00
& 0.00
& 0.00
& 0.00
& 0.00
& 14.26
& 2.000 \\

\midrule

\textbf{VIGOR}
& \cellcolor{bestblue}\textbf{0.30}
& \cellcolor{bestblue}\textbf{99.51}
& \cellcolor{bestblue}\textbf{96.86}
& \cellcolor{bestblue}\textbf{72.39}
& \cellcolor{bestblue}\textbf{59.05}
& \cellcolor{bestblue}\textbf{72.75}
& \cellcolor{bestblue}\textbf{72.03}
& \cellcolor{bestblue}\textbf{66.90}
& \cellcolor{bestblue}\textbf{91.20}
& \cellcolor{bestblue}\textbf{4.050} \\

\bottomrule
\end{tabular}%
}

\end{table*}

\subsection{Ablation Study}

Table~\ref{tab:ablation_results} analyzes the contribution of each component in VIGOR. VIGOR-SFT substantially improves Geometry from 24.78 to 92.79 and Overall from 8.95 to 69.34, but its Feasibility remains only 14.57, indicating that supervised fine-tuning alone cannot ensure complete engineering constraint satisfaction. Applying the refinement pipeline directly to the untuned Base model improves Geometry to 68.42, but Feasibility remains 0.00, showing that refinement alone cannot compensate for the lack of a structured initial program. Starting from the SFT-generated program, local refinement $\mathcal{L}$ raises Feasibility to 56.86 and Overall to 81.18, while joint refinement $\mathcal{J}$ further improves them to 81.43 and 87.11, respectively. Combining both refinement stages yields the full VIGOR, achieving 99.59 Geometry, 96.86 Feasibility, and 91.20 Overall. These results indicate that SFT and verification-guided refinement play complementary roles: SFT provides a feasible initialization space, while $\mathcal{L}$ and $\mathcal{J}$ progressively resolve remaining geometric violations and coupled conflicts.

\begin{table*}[t]
\centering
\caption{Ablation study of different VIGOR variants. Best results are highlighted in blue.}
\label{tab:ablation_results}

\scriptsize
\renewcommand{\arraystretch}{1.08}

\begin{tabular*}{\textwidth}{
@{\extracolsep{\fill}}
l c c c c c
@{}
}
\toprule

\textbf{Method} &
\textbf{Geometry} $\uparrow$ &
\textbf{Functional} $\uparrow$ &
\textbf{Feasibility} $\uparrow$ &
\textbf{Overall} $\uparrow$ &
\textbf{LLMJudge} $\uparrow$ \\

\midrule

Base
& 24.78
& 11.02
& 0.00
& 8.95
& 1.23 \\

Base + $\mathcal{L}\rightarrow\mathcal{J}$
& 68.42
& 22.11
& 0.00
& 24.03
& 1.98 \\

VIGOR-SFT
& 92.79
& 70.31
& 14.57
& 69.34
& \cellcolor{bestblue}\textbf{4.15} \\

VIGOR-SFT + $\mathcal{L}$
& 97.76
& \cellcolor{bestblue}\textbf{70.37}
& 56.86
& 81.18
& 4.10 \\

VIGOR-SFT + $\mathcal{J}$
& 98.88
& 68.40
& 81.43
& 87.11
& 4.13 \\

\textbf{VIGOR}
& \cellcolor{bestblue}\textbf{99.59}
& 68.62
& \cellcolor{bestblue}\textbf{96.86}
& \cellcolor{bestblue}\textbf{91.20}
& 4.05 \\

\bottomrule
\end{tabular*}

\end{table*}

\subsection{Case Study and Error Analysis}

\begin{figure}[H]
    \centering
    \includegraphics[width=\linewidth]{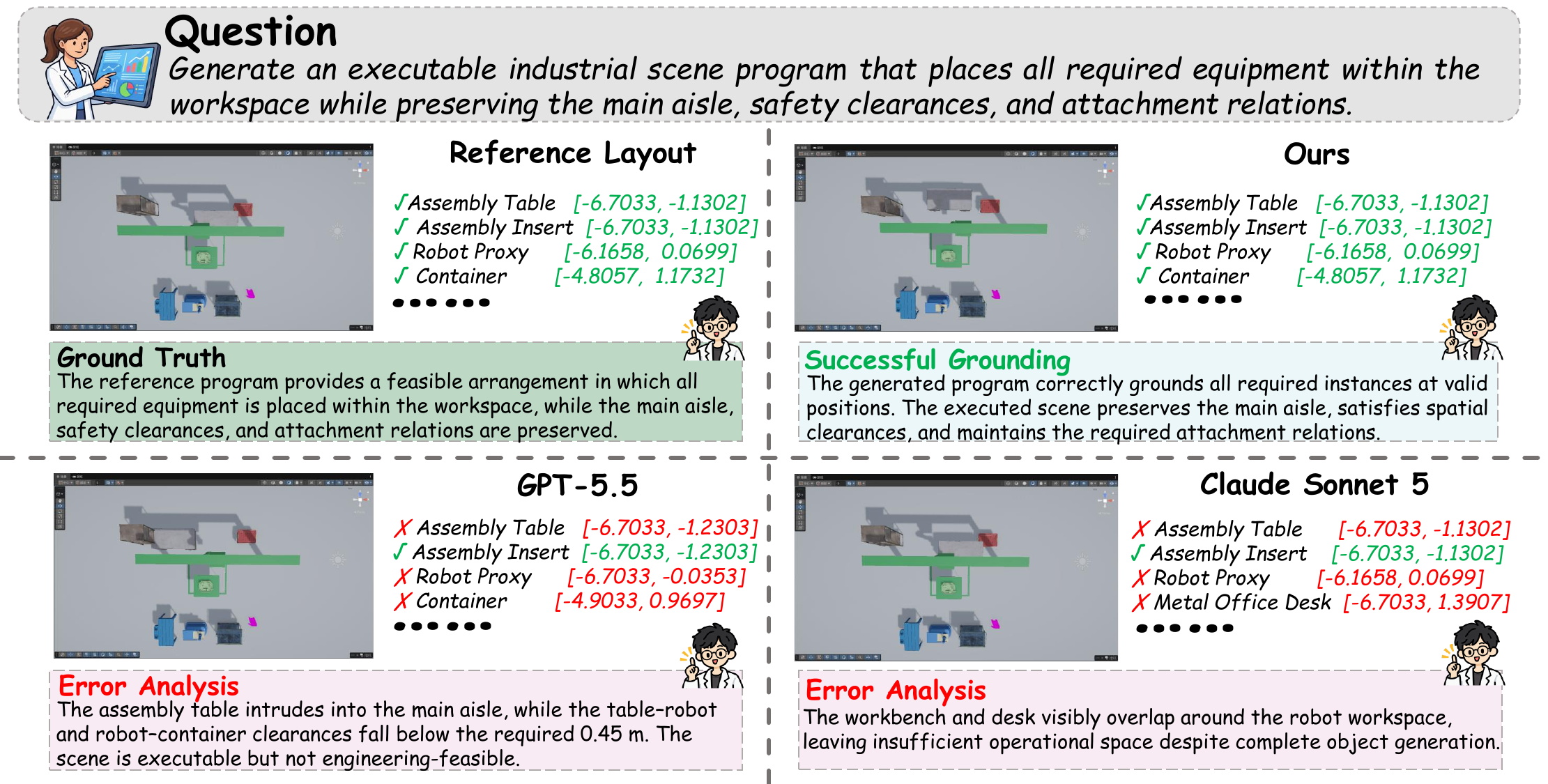}
    \caption{Qualitative comparison. VIGOR better preserves spatial constraints and structural relations, avoiding aisle, clearance, and overlap violations.}
    \label{fig:case}
\end{figure}

Figure~\ref{fig:case} presents a representative industrial scene-program generation task requiring simultaneous satisfaction of aisle, clearance, boundary, and structural constraints. VIGOR produces a feasible layout despite minor numerical deviations. In contrast, GPT-5.5 violates the main aisle and clearance requirements, while Claude Sonnet 5 produces overlapping equipment with insufficient operational space. These failures show that the main challenge lies in coordinating coupled multi-object constraints rather than simply generating complete scene programs. This case further supports the effectiveness of VIGOR's post-execution verification and local--joint geometric refinement.

\subsection{Consistency with Human Judgments}

\begingroup
\setlength{\columnsep}{4pt}

\begin{wrapfigure}[13]{r}{0.35\linewidth}
    \centering
    \vspace{-8pt}
    \includegraphics[width=0.85\linewidth]{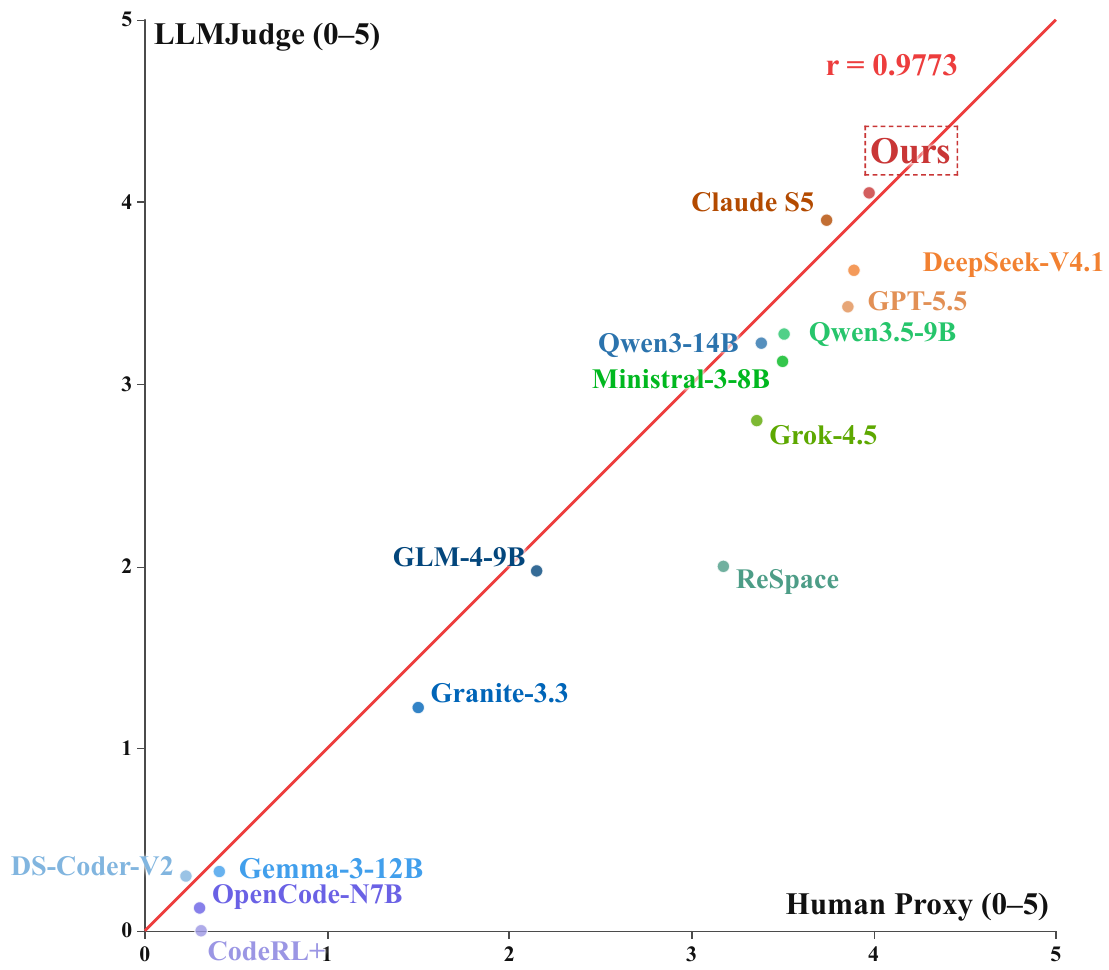}
    \vspace{-4pt}
    \caption{Automatic vs.\ human ratings.}
    \label{fig:human}
    \vspace{-6pt}
\end{wrapfigure}

Figure~\ref{fig:human} shows a clear alignment between LLMJudge and human ratings. Lower-performing methods cluster toward the lower-left region, while stronger models move consistently toward the upper-right. VIGOR occupies the top-right region, achieving high scores under both evaluation protocols. The near-perfect correlation ($r=0.9773$) demonstrates strong agreement between LLMJudge and independent human judgments across the evaluated methods. This consistency indicates that the automated visual evaluation captures model-level differences that are also reflected in human assessments, supporting its use as a reliable complement to the objective engineering metrics. Detailed definitions and aggregation rules for the visual evaluation are provided in Appendix~\ref{app:evaluation_metrics}, and the complete machine-evaluation prompts are provided in Appendix~\ref{app:machine_evaluation_prompts}.

\par
\endgroup

\section{Conclusion}

We study structured industrial scene-program generation from the perspective of complete engineering feasibility, where producing an executable program is not sufficient unless the executed scene also satisfies the required spatial, structural, and functional constraints. This setting is challenging because multiple constraints are coupled through shared device transforms, so resolving one violation may introduce another. To make this problem measurable, we construct an execution-grounded benchmark in which complete \texttt{scene\_program.v1} programs are executed, save/reloaded, and verified against the final scene state. We further propose VIGOR, which combines supervised scene-program generation with post-execution verification and deterministic local--joint geometric refinement. Experiments reveal a clear \emph{executability--engineering-feasibility gap}: strong models can generate structured and geometrically plausible programs while still failing complete engineering feasibility, whereas VIGOR substantially narrows this gap and achieves the strongest overall engineering performance. This work focuses on a controlled industrial layout setting with a fixed asset inventory, structured program schema, and deterministic state-based verification. Extending the same principle to broader asset libraries, richer scene programs, dynamic process constraints, and live simulation or editor environments remains a natural direction for future work.

\subsection*{AI Use Statement}

Generative AI tools were used solely during dataset construction to assist with the generation and organization of structured task instances. The authors designed the benchmark, dataset construction procedures, evaluation criteria, and experimental protocols, and independently verified all generated data and reported results.
\subsection*{Reproducibility Statement}
We provide detailed descriptions of the benchmark construction, 
scene-program specification, execution protocol, and evaluation criteria. 
The paper and appendix include the complete task construction procedure, 
dataset statistics, program interface, validation pipeline, and metric 
definitions. We further report the model architecture, fine-tuning strategy, 
training schedule, inference protocol, and baseline evaluation settings to 
facilitate reproduction of the reported experiments. 

\bibliography{iclr2027_conference}
\bibliographystyle{iclr2027_conference}

\appendix

\section{Performance Results with Radar Chart Analysis} \label{app:capability_profiles}

\begin{figure*}[t]
    \centering
    \includegraphics[width=0.95\textwidth]{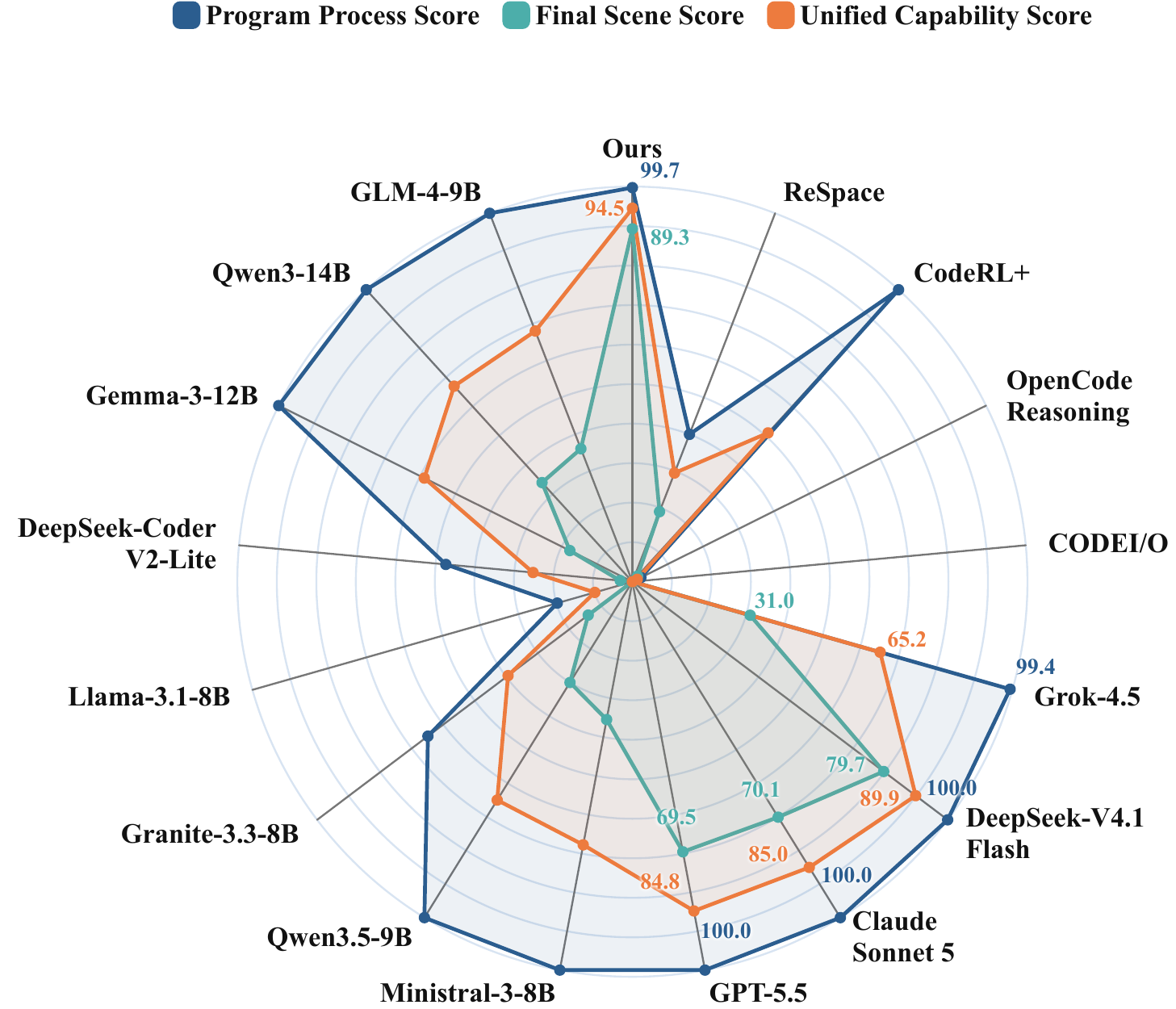}
    \caption{Process--result capability profiles across industrial scene-program
    generation methods. Program Process Score measures structured program
    construction, Final Scene Score measures executed-scene quality, and Unified
    Capability Score summarizes both perspectives.}
    \label{fig:radar}
\end{figure*}

To further analyze the capability profiles of different methods, we compare
VIGOR with representative open-weight models, closed-source/API models, and
published code/layout generation methods across three complementary dimensions,
as shown in Figure~\ref{fig:radar}. \textbf{Program Process Score} measures the
quality of the structured program-construction process, \textbf{Final Scene
Score} evaluates the engineering quality of the executed scene, and
\textbf{Unified Capability Score} summarizes the two perspectives into a single
diagnostic view.

The radar chart reveals a clear process--result mismatch in existing methods.
Several general models, including GLM-4-9B, Qwen3-14B, Gemma-3-12B, Qwen3.5-9B,
and Ministral-3-8B, achieve near-saturated Program Process scores while showing
substantially weaker Final Scene performance. This indicates that reliably
producing a structurally complete program does not necessarily translate into
an engineering-feasible executed scene. The gap is even more pronounced for
CodeRL+, which attains a Program Process score of 100.0 but only 2.0 in Final
Scene quality, illustrating that program-level correctness alone is
insufficient under coupled geometric constraints.

Closed-source models exhibit stronger final-scene capability. GPT-5.5 and
Claude Sonnet 5 reach Final Scene scores of 69.5 and 70.1, respectively, while
DeepSeek-V4.1-Flash further improves this score to 79.7. Nevertheless, their
process and result profiles remain imbalanced, suggesting that high-quality
scene-program generation still requires stronger alignment between structured
program formation and final engineering feasibility.

VIGOR achieves the most balanced capability profile, reaching 99.7 in Program
Process, 89.3 in Final Scene, and 94.5 in Unified Capability. Unlike methods
whose process scores remain high while final-scene quality degrades, VIGOR
maintains strong performance across both stages. These results provide further
evidence of the \emph{executability--engineering-feasibility gap}: successful
program construction is only a prerequisite, while reliable industrial scene
generation additionally requires execution-grounded verification and
coordinated resolution of residual geometric conflicts.

\section{Fine-Grained Task Capability Analysis} \label{app:task_capability}

\begin{figure*}[t]
    \centering
    \includegraphics[width=\textwidth]{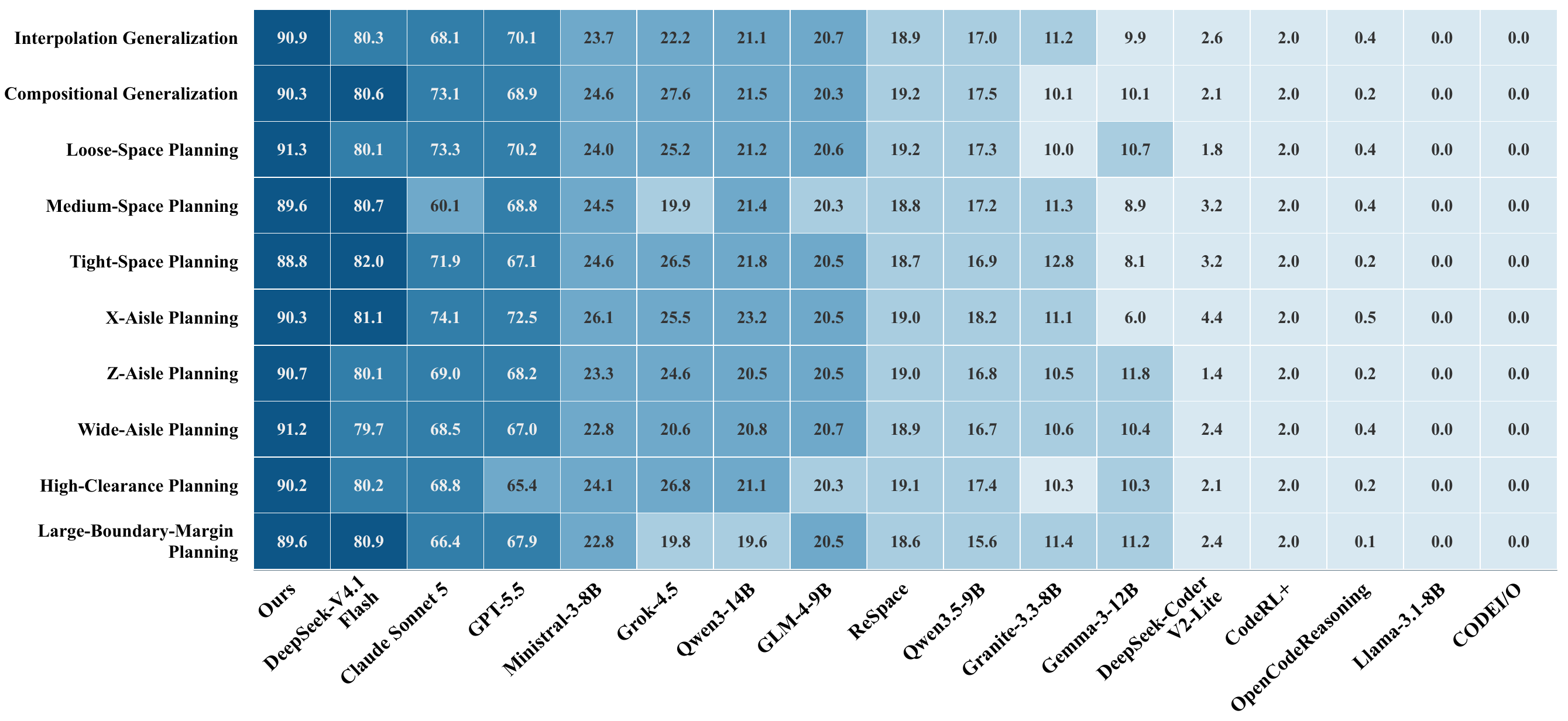}
    \caption{Fine-grained task capability analysis across ten industrial
    scene-generation conditions. Higher values indicate stronger unified
    structural, geometric, and functional capability.}
    \label{fig:capability_heatmap}
\end{figure*}

To further investigate the task-level bottlenecks of existing methods,
we decompose the evaluation into ten capability categories spanning
generalization, constraint density, aisle orientation, and
constraint stress. As shown in Figure~\ref{fig:capability_heatmap},
each cell reports the mean unified task capability over the
corresponding evaluation subset, jointly reflecting structural,
geometric, and functional scene quality.

Across the two evaluation splits, VIGOR maintains consistently high capability, achieving 90.9 on the interpolation split and 90.3 on the held-out compositional split.In contrast, the strongest baseline,
DeepSeek-V4.1-Flash, reaches 80.3 and 80.6, while most open-weight
models remain substantially lower. The small difference between VIGOR's interpolation and compositional scores indicates that its scene-program generation and refinement pipeline maintains strong performance on held-out combinations of engineering conditions, even when those specific parameter combinations are absent from training.

Constraint density exposes a stronger distinction between program
generation and complete scene feasibility. VIGOR achieves 91.3,
89.6, and 88.8 under loose-, medium-, and tight-space planning,
respectively, showing only limited degradation as geometric margins
become more restrictive. DeepSeek-V4.1-Flash remains the strongest
baseline across these settings, whereas several other models exhibit
larger fluctuations. These results suggest that tightly coupled
boundary, collision, clearance, and aisle constraints remain difficult
for general-purpose models even when they can produce structurally
valid scene programs.

A similar pattern appears across aisle orientations. VIGOR reaches
90.3 on X-aisle planning and 90.7 on Z-aisle planning, indicating
little sensitivity to orientation changes. Strong closed-source
models also remain comparatively stable, but at lower absolute
capability levels. In contrast, several weaker models show larger
orientation-dependent variation, suggesting that their generated
layouts are less robust to changes in the spatial organization of the
workspace.

The three stress categories further evaluate challenging input
conditions involving wide aisles, high equipment-clearance
requirements, and large boundary margins. VIGOR scores 91.2, 90.2,
and 89.6 across these settings, respectively, while
DeepSeek-V4.1-Flash remains around 80. These categories place stronger
pressure on coordinated multi-object placement because satisfying one
constraint can reduce the feasible region available to neighboring
devices. The consistently high VIGOR scores therefore indicate that
local correction and joint conflict resolution remain effective even
when the available placement space is constrained by multiple
simultaneous engineering requirements.

Overall, VIGOR exhibits the most stable profile across all ten
capability categories, ranging only from 88.8 to 91.3. This contrasts
with the substantially lower and more variable profiles of existing
general models and published code/layout methods. The fine-grained
results reinforce the central observation of our main experiments:
reliable industrial scene generation requires not only successful
structured-program construction, but also robust final-state
verification and coordinated resolution of coupled engineering
constraints.
\section{Task Construction and Constructibility Screening}
\label{app:task_construction}

Our dataset is constructed from explicit industrial layout conditions.
Each task uses a fixed $7.5 \times 7.0$\,m workspace and eleven
required instances organized into nine top-level equipment groups
and two attachments. Task variation is controlled by the condition
tuple $(w,c,b,o)$, where $w$ is the minimum main-aisle width,
$c$ is the clearance between independent equipment groups,
$b$ is the work-area boundary clearance, and
$o\in\{X,Z\}$ specifies the aisle orientation.

\paragraph{Candidate conditions.}
We enumerate the parameter grid in
Table~\ref{tab:construction_grid}, yielding
$17\times9\times9\times2=2{,}754$ candidate conditions.
Distances are specified in metres, with a uniform increment of
$0.05$\,m. The equipment inventory, asset geometry, workspace,
and required attachment relations remain fixed across conditions.

\begin{table}[h]
\centering
\small
\caption{Parameter grid used for task construction.}
\label{tab:construction_grid}
\begin{tabular}{lcc}
\toprule
Parameter & Values & Count \\
\midrule
Main-aisle width $w$ & $1.00$--$1.80$\,m & 17 \\
Equipment clearance $c$ & $0.20$--$0.60$\,m & 9 \\
Boundary clearance $b$ & $0.30$--$0.70$\,m & 9 \\
Aisle orientation $o$ & $X$, $Z$ & 2 \\
\bottomrule
\end{tabular}
\end{table}

\paragraph{Constructibility screening.}
Screening is performed by a deterministic layout generator and
validator, rather than a language-model classifier.
For each condition, the generator attempts its two predefined
construction variants in a fixed order. A condition is retained
when a generated layout passes the corresponding validation
checks. This procedure retains $1{,}749$ conditions and records
$1{,}005$ unsuccessful conditions.
An unsuccessful construction indicates that the prescribed
procedure did not produce a valid layout; it does not establish
that the condition is mathematically infeasible.The complete condition grid, constructibility-screening protocol, and reference-construction details are provided in Appendix~\ref{app:task_construction}.

\paragraph{Reference construction.}
For each retained condition, the reference pipeline constructs
a baseline layout, performs the prescribed layered candidate
search, and selects the highest-ranked legal candidate.
The selected layout is converted into a complete
\texttt{scene\_program.v1} program without changing its transforms
or attachment relations.
Each condition contributes one reference program. Reference
layouts, execution traces, and verification results are retained
separately from the model-visible task input.

\paragraph{Dataset partition.}
The final partition contains $1{,}399$ training conditions,
$175$ interpolation-test conditions, and $175$
compositional-test conditions.
The compositional split holds out the joint region
$w\in[1.20,1.30]$ and $c\in[0.35,0.55]$ across both aisle
orientations. These width--clearance combinations are absent
from training, while their individual parameter values remain
represented in training for the same orientation.
The interpolation split is selected from the remaining pool
using stratified systematic sampling with seed $42$.
Each interpolation condition has a same-orientation training
neighbour one $0.05$\,m grid step away.
Split assignment does not use model predictions or evaluation
scores.

\section{Scene-Program Specification and Generation Input}
\label{app:program_specification}

The model receives an answer-free task description containing
the public coordinate convention, task constraints, asset
geometry, and program interface.
Condition-dependent fields specify the clearance thresholds,
workspace bounds, and reserved aisle.
The asset catalog provides local collision AABBs and canonical
scales, rather than target object positions.
Reference programs, reference layouts, execution traces, and
evaluation scores are excluded from the input.

\paragraph{Deterministic task template.}
The following box presents the model-visible input in
parameterized form. Braced fields are expanded deterministically
from public task metadata.
The asset-geometry field expands into eleven entries, each
containing the asset name, local AABB minimum and maximum, and
canonical scale. It contains no reference-layout coordinates.

\begin{promptbox}[Deterministic industrial task template]
\begin{lstlisting}
# Code task input (model-visible)

Write one complete scene_program.v1 JSON program for
the supplied public asset catalog.

The program must create nine top-level equipment
groups and the two required attachments, set each
explicit world Transform, establish the requested
parent links, save a new scene, and independently
export object identity, world/local Transform, and
parent relations.

The required output is this program JSON itself;
do not output a layout JSON or a prose-only plan
in its place. Do not load a scene by task number
or use a standard-answer shortcut.

## Public world contract

- Units: metres.
- Coordinate system: Unity left-handed, +Y up,
  X-Z ground plane, +Z forward.
- Position is world position in metres; rotation
  is Unity Euler degrees (x, y, z); scale is
  component-wise.
- Asset root/pivot is the imported prefab root.
  The public normalized assets use the documented
  collision-footprint/lowest-Y convention;
  imported PolyHaven prefabs retain their project
  prefab root and must be verified in Unity.
- Use explicit world transforms and do not silently
  convert, snap, repair, or infer coordinates.

## Public task constraints

- Main aisle minimum width: {w} m;
  aisle orientation: {o}.
- Independent equipment horizontal clearance:
  {c} m.
- Work-area boundary clearance: {b} m.
- Required parent links:
  kit_tray -> AGV at tray_mount;
  assembly_insert -> assembly_table at table_top;
  all other listed devices are top-level with
  ground_anchor.
- The following selected work area and reserved
  aisle are public scene constraints; they do not
  specify an answer layout.
- Selected work area Unity X-Z:
  min={workspace_min} max={workspace_max}.
- Reserved main aisle Unity X-Z:
  min={aisle_min} max={aisle_max}.

## Public asset geometry
## (collision local AABB, metres)

{public_asset_geometry_entries}

Return one JSON object with
schema_version=scene_program.v1 and a complete
ordered operations array. Do not include a save
path, run id, source path, seed id, or
expected-state field.

## JSON program interface (model-visible)

- Allowed operations are exactly create_object,
  set_transform, set_parent, and save_scene.

- create_object:
  {"op":"create_object",
   "instance_id":"...",
   "asset_id":"..."}
  It starts an unparented identity object.

- set_transform:
  {"op":"set_transform",
   "instance_id":"...",
   "position":[x,y,z],
   "rotation_deg":[x,y,z],
   "scale":[sx,sy,sz]}
  Position is world metres and all values must
  be finite; scale is positive.

- set_parent:
  {"op":"set_parent",
   "child_id":"...",
   "parent_id":"...",
   "keep_world_transform":true}
  It keeps the child's world Transform and uses
  full parent rotation/scale conversion.

- All creates come first, then all transforms,
  then parent links, then exactly one final
  save_scene operation. No operation may
  follow save.

- The executor supplies the output path.
  Do not add fields, paths, answer references,
  implicit defaults, coordinate repairs,
  or a solver.

Generic format example
(illustrative identifiers and values only;
not a task answer):

{
  "schema_version": "scene_program.v1",
  "operations": [
    {
      "op": "create_object",
      "instance_id": "device_A",
      "asset_id": "public_asset_A"
    },
    {
      "op": "set_transform",
      "instance_id": "device_A",
      "position": [1.0, 0.0, 2.0],
      "rotation_deg": [0.0, 90.0, 0.0],
      "scale": [1.0, 1.0, 1.0]
    },
    {"op": "save_scene"}
  ]
}
\end{lstlisting}
\end{promptbox}

\paragraph{Program structure.}
A complete reference program contains $25$ operations:
eleven \texttt{create\_object} operations, eleven
\texttt{set\_transform} operations, two
\texttt{set\_parent} operations, and one final
\texttt{save\_scene} operation.
The required instances comprise a kit-tray station, an AGV,
a kit tray, an assembly table, an assembly insert, a ceiling-robot
proxy, a tool cart, a metal office desk, a storage rack,
a container, and a warning sign.
The kit tray is attached to the AGV, and the assembly insert
is attached to the assembly table.

\paragraph{Execution semantics.}
Execution begins from an empty device state.
Each instance must be created once and receive an explicit
world transform before parenting.
Parenting preserves the child's world transform and derives
its local transform using the parent's full translation,
rotation, and scale.
Unknown operations or assets, duplicate identifiers, invalid
numbers, non-positive scales, cyclic parent relations, and
operation-order violations are rejected.
The output path is supplied by the runner rather than generated
by the model.

\section{Execution-Based Dataset Quality Assurance}
\label{app:dataset_quality}

Reference programs are validated through execution and
independent state checks.
A valid JSON representation alone is insufficient:
the program must produce the required scene structure and
satisfy the public geometric constraints after execution.
Dataset acceptance uses deterministic checks rather than
language-model ratings.

\paragraph{Schema and execution checks.}
Each program is checked against the
\texttt{scene\_program.v1} schema and executed by the Python
state engine in operation order.
The execution trace records the resulting state as operations
are applied.
A successful run must reach the final
\texttt{save\_scene} operation and produce a new state file.
Diagnostic partial states from failed execution are not accepted
as completed scenes.

\paragraph{Identity, transforms, and attachments.}
The saved state must contain all eleven required instances
with unique identities and the correct asset assignments.
World and local transforms, parent links, and placement anchors
are checked against the reference construction and public task
contract.
The two required attachments must be preserved, while the
remaining equipment forms nine top-level groups.
Reference information is used for post-execution verification;
it is not supplied to the executor to complete or repair
the program.

\paragraph{Independent geometry verification.}
World collision AABBs are computed by transforming the eight
corners of each public local AABB using the executed transform.
The resulting geometry is independently recomputed and checked
for consistency.
The unified geometric evaluation applies $94$ checks per scene,
covering boundary clearance, main-aisle preservation,
pairwise collision avoidance, and independent-equipment
clearance.
Structural and attachment checks are evaluated in addition
to these geometric checks.

\paragraph{Save--reload consistency.}
The runner independently reloads the newly saved state and
checks that its serialized object state is preserved.
This verifies persistence of the Python scene representation,
including identities, transforms, and parent relations.
It does not constitute verification of Unity prefab identity,
Unity collider geometry, or Unity scene save--reload behaviour.

\paragraph{Input isolation.}
The model-visible input is checked against the fixed task
template. Only five public-constraint lines vary across
conditions: the main-aisle requirement, equipment clearance,
boundary clearance, workspace bounds, and reserved-aisle bounds.
Additional checks inspect source identifiers, file paths,
expected-state references, and reference-position triples.
Test inputs and reference artifacts are stored separately;
test-request construction reads the answer-free inputs.

\paragraph{Validation outcome.}
All $1{,}749$ retained conditions pass program conversion,
schema validation, state-engine execution, save--reload
verification, identity and attachment checks, independent
geometry verification, and the unified hard-constraint
evaluation.
The resulting reference set therefore contains
$1{,}749\times94=164{,}406$ passing geometric checks.
These results establish consistency under the public asset
geometry and Python execution semantics; they do not imply
that every reference scene has been executed in Unity.

\section{Machine Evaluation Prompts}
\label{app:machine_evaluation_prompts}

We use \texttt{gpt-5.6-terra} to evaluate each valid anonymous
scene card. Operational Usability (OU) and Spatial Organization
(SO) are evaluated in separate calls. The judge returns an
integer score from 0 to 5 and a concise evidence-based reason
for one dimension.

\begin{promptbox}[Machine judge prompt for Operational Usability]
\begin{lstlisting}
You are an expert industrial-engineering assessor. You are judging ONE dimension of ONE candidate workcell layout: **OPERATIONAL USABILITY (OU)**.

You are shown a single anonymous evaluation card. It contains:
* a plan view (X-Z) of the final scene produced by executing one candidate program, with the figure legend that explains every colour and marker;
* the public task condition of the question;
* the work area and the reserved main aisle, which are public scene constraints and not an answer;
* the eleven devices and the function each one plays;
* the executed world transform of every device in the scene.

## What you judge

Judge only how usable the resulting workcell would be in real industrial operation:

* whether the devices are convenient to operate and to maintain;
* whether material handling and the human/robot workflow are sensible;
* whether devices that have to work together are placed so that they can work together;
* whether the arrangement contains something that would visibly get in the way of real use.

## What you must not do

* **Do not re-derive centimetre-level geometry.** Do not measure exact distances, do not recompute boundary margins, net clearances or collisions, and do not check whether the reserved aisle band is exactly wide enough. A separate automatic geometric scorer already owns those exact checks, and they cannot be reproduced reliably from a figure.
* **Do not compare the layout against a reference arrangement.** The task admits many legal solutions. Never penalise a layout for not resembling some imagined answer.
* **Do not try to guess which method produced this layout**, and do not let any such guess affect the score.

## Scale (integer 0-5)

5  professional and sensible; almost nothing would need changing
4  good overall; only minor shortcomings
3  basically usable, but with clearly unreasonable aspects
2  several important problems; hard to use in practice
1  most of the arrangement is not sensible
0  no valid scene, or completely unusable

## Output

Reply with exactly one JSON object and nothing else:

{"anonymous_id": "<the anonymous_sample_id printed on the card>", "dimension": "OU", "score": 4, "reason": "Concise evidence-based explanation."}

score must be an integer between 0 and 5. reason must be one or two sentences naming the evidence you actually saw on the card. Do not output a total score, and do not output any field other than these four.
\end{lstlisting}
\end{promptbox}

\begin{promptbox}[Machine judge prompt for Spatial Organization]
\begin{lstlisting}
You are an expert industrial-engineering assessor. You are judging ONE dimension of ONE candidate workcell layout: **SPATIAL ORGANIZATION (SO)**.

You are shown a single anonymous evaluation card. It contains:
* a plan view (X-Z) of the final scene produced by executing one candidate program, with the figure legend that explains every colour and marker;
* the public task condition of the question;
* the work area and the reserved main aisle, which are public scene constraints and not an answer;
* the eleven devices and the function each one plays;
* the executed world transform of every device in the scene.

## What you judge

Judge only how well the layout is organised in space:

* whether the functional zones are clear;
* whether related devices are sensibly grouped together;
* whether the devices are oriented in a coordinated way rather than at odds with each other;
* whether the space is used compactly without becoming excessively crowded;
* whether the overall arrangement reads as clear and orderly.

## What you must not do

* **Do not re-derive centimetre-level geometry.** Do not measure exact distances, do not recompute boundary margins, net clearances or collisions, and do not check whether the reserved aisle band is exactly wide enough. A separate automatic geometric scorer already owns those exact checks, and they cannot be reproduced reliably from a figure.
* **Do not compare the layout against a reference arrangement.** The task admits many legal solutions. Never penalise a layout for not resembling some imagined answer.
* **Do not try to guess which method produced this layout**, and do not let any such guess affect the score.

## Scale (integer 0-5)

5  professional and sensible; almost nothing would need changing
4  good overall; only minor shortcomings
3  basically usable, but with clearly unreasonable aspects
2  several important problems; hard to use in practice
1  most of the arrangement is not sensible
0  no valid scene, or completely unusable

## Output

Reply with exactly one JSON object and nothing else:

{"anonymous_id": "<the anonymous_sample_id printed on the card>", "dimension": "SO", "score": 4, "reason": "Concise evidence-based explanation."}

score must be an integer between 0 and 5. reason must be one or two sentences naming the evidence you actually saw on the card. Do not output a total score, and do not output any field other than these four.
\end{lstlisting}
\end{promptbox}

```latex
\section{Evaluation Metrics}
\label{app:evaluation_metrics}

We evaluate each method from three complementary perspectives:
rule-based engineering metrics computed from executed scene states,
visual judgments of scene quality, and an aggregate engineering
score. Unless otherwise specified, metrics are computed using the
method's actual evaluation set. Invalid or unscorable outputs remain
in the denominator and receive the prescribed zero score.

\subsection{Structural and Geometric Metrics}

\paragraph{Parent--Attach.}
Each task requires the specified parent relations and attachment
relations to be satisfied. Let $r_i^{\mathrm{parent}}$ and
$r_i^{\mathrm{attach}}$ denote the corresponding ratios for task
$i$. We count a task as structurally correct only when both relations
are fully satisfied:
\begin{equation}
z_i^{\mathrm{PA}}
=
\mathbb{I}
\left[
r_i^{\mathrm{parent}}=1
\land
r_i^{\mathrm{attach}}=1
\right],
\qquad
\mathrm{Parent\text{-}Attach}
=
\frac{1}{N}\sum_{i=1}^{N} z_i^{\mathrm{PA}} .
\label{eq:parent_attach}
\end{equation}
The reported value lies in $[0,1]$. A task without a valid scene
state receives zero.

\paragraph{Geometric Violation Burden.}
The geometric verifier evaluates 94 constraint atoms per task:
11 boundary-clearance atoms, 36 pairwise collision atoms, 36
pairwise equipment-clearance atoms, and 11 main-aisle atoms.
Let $u_{i,k}$ denote the number of unsatisfied atoms in constraint
family $k$ for task $i$. The per-task and method-level violation
burden are
\begin{equation}
\begin{aligned}
v_i &=
u_{i,\mathrm{boundary}}
+u_{i,\mathrm{collision}}
+u_{i,\mathrm{clearance}}
+u_{i,\mathrm{aisle}},\\
\mathrm{GVB} &= \frac{1}{N}\sum_{i=1}^{N}v_i .
\end{aligned}
\label{eq:gvb}
\end{equation}
GVB ranges from 0 to 94, with lower values indicating fewer
unsatisfied geometric constraints. A task without a valid scene
state is assigned $v_i=94$.

\paragraph{Bottleneck Geometry.}
GVB counts violated atoms jointly, whereas Bottleneck Geometry
measures the weakest of the four geometric constraint families in
each task. For family $k$, let $n_k$ be its number of atoms and
define its satisfaction rate as
\begin{equation}
\rho_{i,k}=1-\frac{u_{i,k}}{n_k},
\qquad
(n_{\mathrm{boundary}},n_{\mathrm{collision}},
n_{\mathrm{clearance}},n_{\mathrm{aisle}})
=(11,36,36,11).
\label{eq:geometry_family_rate}
\end{equation}
We then compute
\begin{equation}
\mathrm{BottleneckGeometry}
=
\frac{1}{N}\sum_{i=1}^{N}
100\min_{k\in\mathcal{K}}\rho_{i,k},
\quad
\mathcal{K}=
\{\mathrm{boundary},\mathrm{collision},
\mathrm{clearance},\mathrm{aisle}\}.
\label{eq:bottleneck_geometry}
\end{equation}
The minimum is taken per task before averaging across tasks.
Therefore, Bottleneck Geometry is not obtained by linearly
converting the mean GVB.

\paragraph{Complete Feasibility.}
Let $f_i$ be the task-level complete-feasibility indicator recorded
by the authoritative evaluator. It is one only when the required
execution and final-state hard constraints pass, and zero otherwise.
The method-level score is
\begin{equation}
\mathrm{CompleteFeasibility}
=
\frac{100}{N}\sum_{i=1}^{N} f_i .
\label{eq:complete_feasibility}
\end{equation}
This metric is reported on a 0--100 scale.

\subsection{Functional Quality and Test-Subset Metrics}

\paragraph{Functional Layout.}
The authoritative evaluator assigns six functional-quality scores
to each task: process-chain organization, tool/kitting organization,
rack placement, container placement, desk placement, and hazard-sign
placement. Let $q_{i,j}\in[0,100]$ denote the score for task $i$ and
functional dimension $j$. The task-level and method-level scores are
\begin{equation}
\mathrm{FL}_i=\frac{1}{6}\sum_{j=1}^{6}q_{i,j},
\qquad
\mathrm{FunctionalLayout}
=\frac{1}{N}\sum_{i=1}^{N}\mathrm{FL}_i .
\label{eq:functional_layout}
\end{equation}
All six dimensions are equally weighted. Invalid or unscorable tasks
receive zero and remain in the denominator.

\paragraph{Functional-quality subsets.}
We report four additional summaries of Functional Layout. BQ-FQ
measures performance on the lower tail: after sorting the $N$
task-level scores in ascending order, it averages the lowest
$\lceil 0.25N\rceil$ scores. Interp-FQ and Comp-FQ are the
full-denominator means over the frozen interpolation and
compositional test subsets, respectively. Tight-FQ is the mean over
the frozen tight-condition subset. For methods evaluated on a
reduced test set, the corresponding frozen subset assignment and
denominator are used; subsets are not selected using model scores.

\subsection{Overall Score}

Overall summarizes structural correctness, geometric quality,
functional quality, and complete feasibility. All four components
are placed on a 0--100 scale before averaging. We define
\begin{equation}
\begin{aligned}
S &= 100\times\mathrm{Parent\text{-}Attach},\\
GS &= 100\left(1-\frac{\mathrm{GVB}}{94}\right),\\
G &= \frac{GS+\mathrm{BottleneckGeometry}}{2},\\
Q &= \frac{
\mathrm{FunctionalLayout}
+\mathrm{BQ\text{-}FQ}
+\mathrm{Interp\text{-}FQ}
+\mathrm{Comp\text{-}FQ}
+\mathrm{Tight\text{-}FQ}
}{5},\\
F &= \mathrm{CompleteFeasibility},\\
\mathrm{Overall} &= \frac{S+G+Q+F}{4}.
\end{aligned}
\label{eq:overall}
\end{equation}
Here, $GS$ is the overall geometric-atom satisfaction rate, while
$G$ combines this rate with the per-task bottleneck measure.
Overall gives equal weight to structural correctness, geometric
quality, functional quality, and complete feasibility. Visual
judgments are reported separately and are not included in Overall.

\subsection{Visual Judgment Scores}

We use 20 fixed dual-view evaluation cards per method. Each card
shows the executed scene and its overview under the same legend.
The visual judges score Operational Usability (OU) and Spatial
Organization (SO) independently on an integer scale from 0 to 5.

\paragraph{LLMJudge.}
A single machine judge evaluates each card in separate OU and SO
calls. For method $m$, let $J_{m,i}^{\mathrm{OU}}$ and
$J_{m,i}^{\mathrm{SO}}$ denote the two scores for card $i$. We
compute
\begin{equation}
\begin{aligned}
J_m^{\mathrm{OU}} &=
\frac{1}{N_{\mathrm{visual}}}
\sum_{i=1}^{N_{\mathrm{visual}}}J_{m,i}^{\mathrm{OU}},\\
J_m^{\mathrm{SO}} &=
\frac{1}{N_{\mathrm{visual}}}
\sum_{i=1}^{N_{\mathrm{visual}}}J_{m,i}^{\mathrm{SO}},\\
\mathrm{LLMJudge}_m &=
\frac{J_m^{\mathrm{OU}}+J_m^{\mathrm{SO}}}{2}.
\end{aligned}
\label{eq:llmjudge}
\end{equation}
LLMJudge is reported on the original 0--5 scale.

\paragraph{Human Rating.}
Human Rating aggregates the scores from three independent human
raters. Each rater independently evaluates the same fixed set of
visual evaluation cards using the same OU and SO criteria.
Let $R$ be the set of the three human raters and
$H_{r,m,i}^{d}$ the score assigned by human rater $r$ to method $m$,
card $i$, and dimension $d\in\{\mathrm{OU},\mathrm{SO}\}$. We first
average over cards and raters for each dimension, then average
the two dimensions:
\begin{equation}
\begin{aligned}
H_m^{d}
&=
\frac{1}{|R|N_{\mathrm{visual}}}
\sum_{r\in R}\sum_{i=1}^{N_{\mathrm{visual}}}H_{r,m,i}^{d},\\
\mathrm{HumanRating}_m
=
\frac{H_m^{\mathrm{OU}}+H_m^{\mathrm{SO}}}{2}.
\end{aligned}
\label{eq:human_rating}
\end{equation}
Human Rating is also reported on the 0--5 scale. LLMJudge and Human
Rating are presented as separate visual-evaluation measures and are
not combined with one another or with Overall.
```

\end{document}